\documentclass{article}
\pdfoutput=1
\usepackage[nonatbib,preprint]{neurips_2020}
\usepackage{graphicx}
\usepackage[utf8]{inputenc} 
\usepackage[T1]{fontenc}    
\usepackage{hyperref}       
\usepackage{url}            
\usepackage{booktabs}       
\usepackage{amsfonts}       
\usepackage{nicefrac}       
\usepackage{microtype}      
\usepackage{cite}
\usepackage{appendix}
\usepackage{multirow}
\usepackage{booktabs}
\usepackage{cleveref}
\title{MindDiffuser: \\Controlled Image Reconstruction from Human Brain Activity with Semantic and Structural Diffusion}
\author{
Yizhuo Lu$^{1,2}$ \thanks{Equal contributions.}\ \ ,\ \
\newcommand\CoAuthorMark{\footnotemark[\arabic{footnote}]}
 Changde Du$^{1}$ \CoAuthorMark \ \ ,\ \
 Dianpeng Wang$^{2}$  ,\ \
 Huiguang He$^{1}$\thanks{Corresponding author: Huiguang He}\\
\texttt{luyizhuo2023@ia.ac.cn, changde.du@ia.ac.cn, wdp@bit.edu.cn}\\
\texttt{huiguang.he@ia.ac.cn}\\
$^1$ Research Center for Brain-Inspired Intelligence, Institute of Automation, CAS \\
$^2$ School of Mathematics and Statistics, Beijing Institute of Technology\\
}

\begin{document}

\maketitle
\begin{abstract}
	Reconstructing visual stimuli from measured functional magnetic resonance imaging (fMRI) has been a meaningful and challenging task. Previous studies have successfully achieved reconstructions with structures similar to the original images, such as the outlines and size of some natural images. However, these reconstructions lack explicit semantic information and are difficult to discern. In recent years, many studies have utilized multi-modal pre-trained models with stronger generative capabilities to reconstruct images that are semantically similar to the original ones. However, these images have uncontrollable structural information such as position and orientation. To address both of the aforementioned issues simultaneously, we propose a two-stage image reconstruction model called MindDiffuser, utilizing Stable Diffusion. In Stage 1, the VQ-VAE latent representations and the CLIP text embeddings decoded from fMRI are put into the image-to-image process of Stable Diffusion, which yields a preliminary image that contains semantic and structural information. In Stage 2, we utilize the low-level CLIP visual features decoded from fMRI as supervisory information, and continually adjust the two features in Stage 1 through backpropagation to align the structural information. The results of both qualitative and quantitative analyses demonstrate that our proposed model has surpassed the current state-of-the-art models in terms of reconstruction results on Natural Scenes Dataset (NSD). Furthermore, the results of ablation experiments indicate that each component of our model is effective for image reconstruction.
	
\end{abstract}
\section{Introduction}
The human visual system possesses the exceptional ability to efficiently and robustly perceive and comprehend complex visual stimuli in the real world, which is unparalleled by current artificial intelligence systems. Studying the functions of different brain regions through neural encoding and decoding models can provide us with a deeper understanding of the human visual perception system. Neural encoding pertains to the procedure of transforming external visual stimuli into neural signals, while neural decoding entails establishing a correspondence from neural signals to their corresponding visual stimuli. Depending on the complexity and specific objectives, the latter can be categorized into stimuli classification, stimuli recognition, and stimuli reconstruction\cite{rakhimberdina2021natural}. Stimuli classification refers to using brain activities to predict the discrete object class of a presented stimulus, while stimuli recognition aims to identify a specific stimulus from a known set of images that corresponds to a given pattern of brain activities. Stimuli reconstruction refers to the direct generation of images from provided brain activities. The generated image should accurately align with the corresponding image stimulus in terms of its shape, position, orientation, and other specific details. Despite its immense difficulty, stimuli reconstruction is a critical step towards the ultimate goal of deciphering the workings of the human brain. With the advancement of sophisticated image reconstruction methods and the increase in the volume of neuroimaging data, researchers are increasingly focusing on this direction.

Recently, researches have revealed that deep learning frameworks exhibit a certain level of consistency with the hierarchical encoding-decoding process of the human visual system\cite{pinto2009high}\cite{krizhevsky2009learning}\cite{schrimpf2018brain}. As a result, numerous studies have extensively employed deep neural networks (DNN) for reconstructing natural images. Based on the structure of the previous image reconstruction models, we categorize them into  optimized models and generative models. The optimized model is represented by DGN\cite{shen2019deep} proposed by Shen et al., which utilizes image features extracted from a DNN as a constraint, and optimizes the latent space of the image generator to achieve similarity with the decoded DNN features. While this method allows for alignment of the position, orientation, and size of the reconstructed images with the corresponding ones in pixel space, the absence of image prior knowledge in latent space means that optimization starting from Gaussian noise can result in indistinct outcomes and a lack of clear semantic information. The generative reconstruction models involve decoding fMRI into the latent space of models such as VAE, GAN, and Diffusion model, and leveraging their powerful generation capabilities to reconstruct images that are semantically similar to the original. While this paradigm enables rapid generation of realistic and semantically rich reconstruction images, the outcomes are always lacking in control over details such as location and size.

Building upon the preceding discussion, we present a two-stage image reconstruction model that resolves issues with the two aforementioned reconstruction paradigms, resulting in semantically similar and structurally aligned reconstruction outcomes.

\begin{figure}[htbp!]
	\centering
	\includegraphics[width=1.0\linewidth]{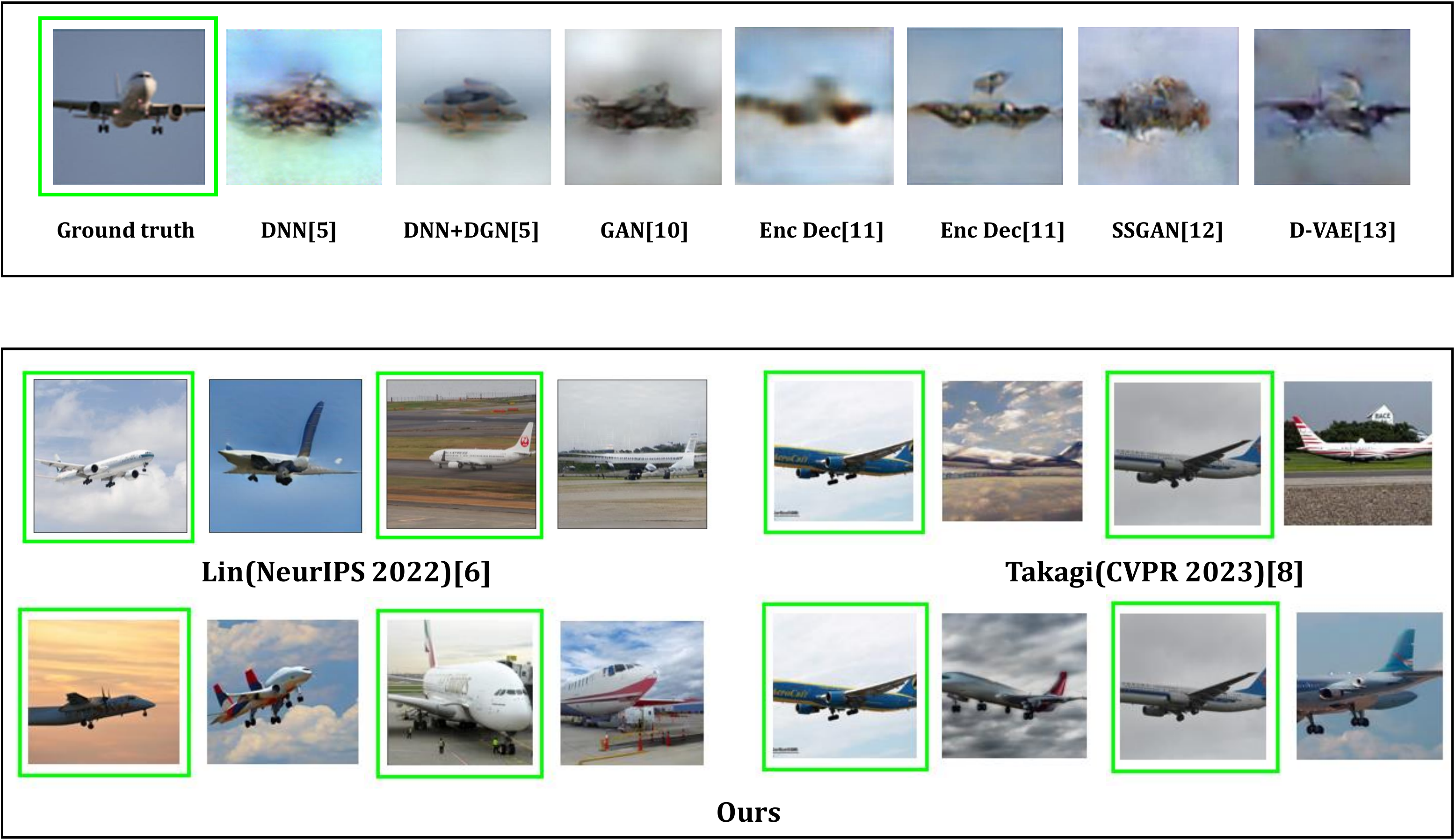}
	\caption{Comparison of image reconstruction results. Note that the subfigure above shows the results of an previous image reconstruction collated by Lin\cite{lin2022mind}.}
	\label{fig:plane}
\end{figure}

Given that various reconstruction models employ distinct datasets, we opt to compare results fairly by utilizing the aircraft reconstruction outcomes, which are prevalent across the majority of datasets and facilitate intuitive comparison. Figure \ref{fig:plane} demonstrates that previous image reconstruction models, such as Shen DNN\cite{shen2019deep} and Shen GAN\cite{shen2019deep}, produce outcomes that resemble the original images in terms of size, shape, and orientation, but lack semantic information necessary for recognition as an airplane. Recent image reconstruction models, such as Mind Reader\cite{lin2022mind} and Takagi's\cite{takagi2022high}, incorporate text representations via multimodal pre-trained models, resulting in reconstructions with correct semantic information. However, in comparison to our MindDiffuser, their reconstructed aircrafts cannot be aligned with the original images on structural information such as shape and posture.

\section{Related works}
\label{related_works}
\subsection{Neural decoding and image reconstruction models}
\label{subsection:related_works_one}
Building upon the pioneering work of Haxby \cite{haxby2001distributed} and other researchers, a multitude of neural decoding tasks with significant guiding implications have surfaced in recent decades. These tasks can be categorized into stimuli classification\cite{haxby2001distributed}\cite{van2010efficient}\cite{damarla2013decoding}\cite{yargholi2016brain}\cite{du2022decoding}, stimuli recognition\cite{haynes2006decoding}\cite{kay2008identifying}\cite{naselaris2009bayesian}\cite{horikawa2017generic}, and stimuli reconstruction\cite{nishimoto2011reconstructing} based on their decoding objectives. Among them, stimuli reconstruction is the most alluring and demanding, and we focus on this area in the study.

Previous image reconstruction techniques utilized linear regression models to fit fMRI with manually defined image features\cite{kay2008naselaris}\cite{naselaris2009bayesian}\cite{fujiwara2013modular}. These approaches predominantly concentrated on extracting pre-established low-level characteristics from image stimuli, such as local image structures or raster filter features. The outcomes obtained from these reconstruction methods were blurry, and the features relied heavily on manual configuration.
With the advent of deep learning, the use of DNNs in this domain has become more prevalent. Shen et al.\cite{shen2019deep} continuously optimized the latent space of the generation model to ensure that the DNN features of the generated images are comparable to the decoding features. Beliy et al.\cite{beliy2019voxels} and Gaziv et al.\cite{gaziv2022self} employed semi-supervised learning\cite{chapelle2009semi} to train an Encoder-Decoder structure to reconstruct images, solving the problem of fMRI-stimulus pairs deficiency. However, the reconstructions of these models only match the original images in terms of their outlines and postures, and do not possess distinguishable semantic information.
Du et al.\cite{du2018reconstructing} introduced a multi-view reconstruction model that accounts for the statistical correlation between fMRI signals and the corresponding stimuli. The model employs Bayesian inference to achieve precise reconstruction outcomes. Du et al.\cite{du2020conditional}\cite{du2020structured} proposed a hierarchical neural decoding framework that leverages structural information and visual features from brain activity. This framework enhances the quality of partial natural images and face stimuli reconstructions through the use of a DNN and a matrix variable Gaussian priori, employing multi-task transfer learning.

Previous researches have mainly focused on reconstructing specific types of stimuli, such as faces or handwritten digits. With the advancement of image generation models and the availability of more adequate brain imaging data, it is now becoming feasible to reconstruct complex natural images. Chen et al.\cite{chen2022seeing} pre-trained fMRI data using a method similar to MAE\cite{he2022masked}, and fine-tuned the LDM\cite{rombach2022high} using the extracted characterizations from the 2D fMRI structure to obtain reconstructed images.
Ozcelik et al.\cite{ozcelik2022reconstruction} and Gu et al.\cite{gu2022decoding} employed a self-supervised model for image instance feature extraction, followed by iterative optimization of noise and dense information using backpropagation. Subsequently, the three features were mapped from fMRI and put into an IC-GAN\cite{casanova2021instance} for image reconstruction.

Since the introduction of multimodal pre-trained model CLIP\cite{radford2021learning}, semantic information from text has been leveraged for reconstructing complex natural images. Lin et al.\cite{lin2022mind} trained a mapping model using contrastive and adversarial learning loss to align fMRI with CLIP latent representations. The mapped fMRI was then fed into StyleGAN2\cite{karras2020analyzing} during the generation stage. Takagi et al.\cite{takagi2022high} achieved close-to-original reconstruction results by mapping fMRI to the text feature $c$ and image feature $z$ of Stable Diffusion\cite{rombach2022high}.
The aforementioned reconstruction models incorporate rich semantic information into the reconstruction results through the direct inclusion of text features or the extraction of features from images. However, since no additional constraints are imposed on the generator during reconstruction process, the quality of the generated image is exclusively determined by the decoded information. Consequently, the resulting reconstruction may lack precision in details such as location, size, and shape.

\subsection{CLIP for image generation}
CLIP\cite{radford2021learning} utilizes image-text contrastive learning to endow its representation space with rich semantic information, which has been widely utilized to guide downstream image generation tasks. Here, We describe the following works according to the different roles that CLIP plays in downstream tasks.

{\bfseries Feed-forward image generation based on CLIP representations:}
OpenAI presents GLIDE\cite{nichol2021glide} for generating images guided by textual descriptions. GLIDE accomplishes this by incorporating text embeddings into the U-net for noise prediction, and employs CLIP-based guidance during the sampling process. DALL-E2\cite{ramesh2022hierarchical} employs a prior model to transform CLIP text embeddings into corresponding image features, which are subsequently utilized by a diffusion model to generate images that are semantically coherent and visually realistic. Stable Diffusion\cite{rombach2022high} utilizes CLIP text embeddings in VQ-VAE's latent space to guide the diffusion model in image generation, thereby enhancing model training and inference speed while ensuring high quality output.

Besides its direct application in guiding image generation, the features extracted by CLIP can also serve as supervisory signals for continuously refining the latent vectors of the image generator via backpropagation. This technique enables fine-grained and personalized image generation and editing.

{\bfseries Backward image optimization supervised by CLIP: }
StyleCLIP\cite{patashnik2021styleclip} begins by obtaining the latent variable for a given face image through GAN Inversion\cite{xia2022gan}. It then optimizes the latent variable under the supervision of the input text's CLIP feature, resulting in an image that aligns with the text in the CLIP representation space. This optimization process enables the generation of images that match the desired style and content specified by the input text. StyleGAN-nada\cite{gal2021stylegan} further excavates the semantic information in the CLIP representation space and enables cross-domain image generation. CLIP-GLaSS\cite{galatolo2021generating} calculates the similarity of CLIP embeddings between the input text and the generated image, and uses a genetic algorithm to generate a image that most closely matches the input. Through the visual branch of CLIP,  CLIPasso\cite{vinker2022clipasso} extracts the structure and semantic information of the original image and the generated stick figure image, whose L2 distance is used as a loss function, and then optimizes the stroke parameters until convergence through backpropagation. Similar to CLIPasso, our MindDiffuser uses the low-level image features extracted from CLIP to constrain the structural information of the reconstructed images.

The prior works leveraging the CLIP representation space mentioned in \ref{subsection:related_works_one} (Lin\cite{lin2022mind}, Takagi\cite{takagi2022high}) utilize the decoded CLIP features to guide the image reconstruction process directly. To the best of our knowledge, our proposed MindDiffuser is the first approach to leverage CLIP features as supervisory information for achieving fine-grained and controllable image reconstruction.

\section{Method}
\subsection{Prerequisites}
{\bfseries Stable Diffusion}: Diffusion models\cite{wijmans1995solution}\cite{ho2020denoising} represent a novel category of probabilistic generative models, whose generative capacities in the realm of computer vision have progressively equaled or even outperformed GANs in certain specialized tasks.
The diffusion models comprise a forward diffusion process and a reverse denoising process, both of which exhibit Markovian properties. In the forward diffusion process, a Gaussian noise is continuously added to the original image until it completely collapses to standard Gaussian noise. The forward diffusion process can be represented as $q(x_t|x_{t-1})=N(x_t;\sqrt{\alpha_t}x_{t-1},(1-\alpha_t)I)$, where t denotes the time step of each noise addition. The reverse denoising process utilizes the U-Net\cite{ronneberger2015u} architecture to effectively fit the noise added at each time step t. Subsequently, the image is generated by sequentially denoising and sampling, commencing from standard Gaussian noise.

In the context of image generation tasks, the conventional diffusion model executes two Markov processes in a large pixel space, resulting in substantial computational resource utilization. To address this issue, Latent Diffusion Models (LDM)\cite{rombach2022high} employs a VQ-VAE\cite{van2017neural} encoder to transform the pixel space into a low-dimensional latent space. Subsequently, the diffusion model's training and generation are performed in the latent space, with the final generated image obtained by utilizing the VQ-VAE decoder. This approach significantly reduces computational resource requirements and inference time while preserving the quality of the generated images. In this paper, we utilize Stable Diffusion to perform image reconstruction tasks. By incorporating a cross-attention mechanism, the CLIP text embeddings were integrated into the U-Net, resulting in images that aligned with the semantic information.

\subsection{Overview}

\begin{figure}[htbp!]
	\centering
	\includegraphics[width=1.0\linewidth]{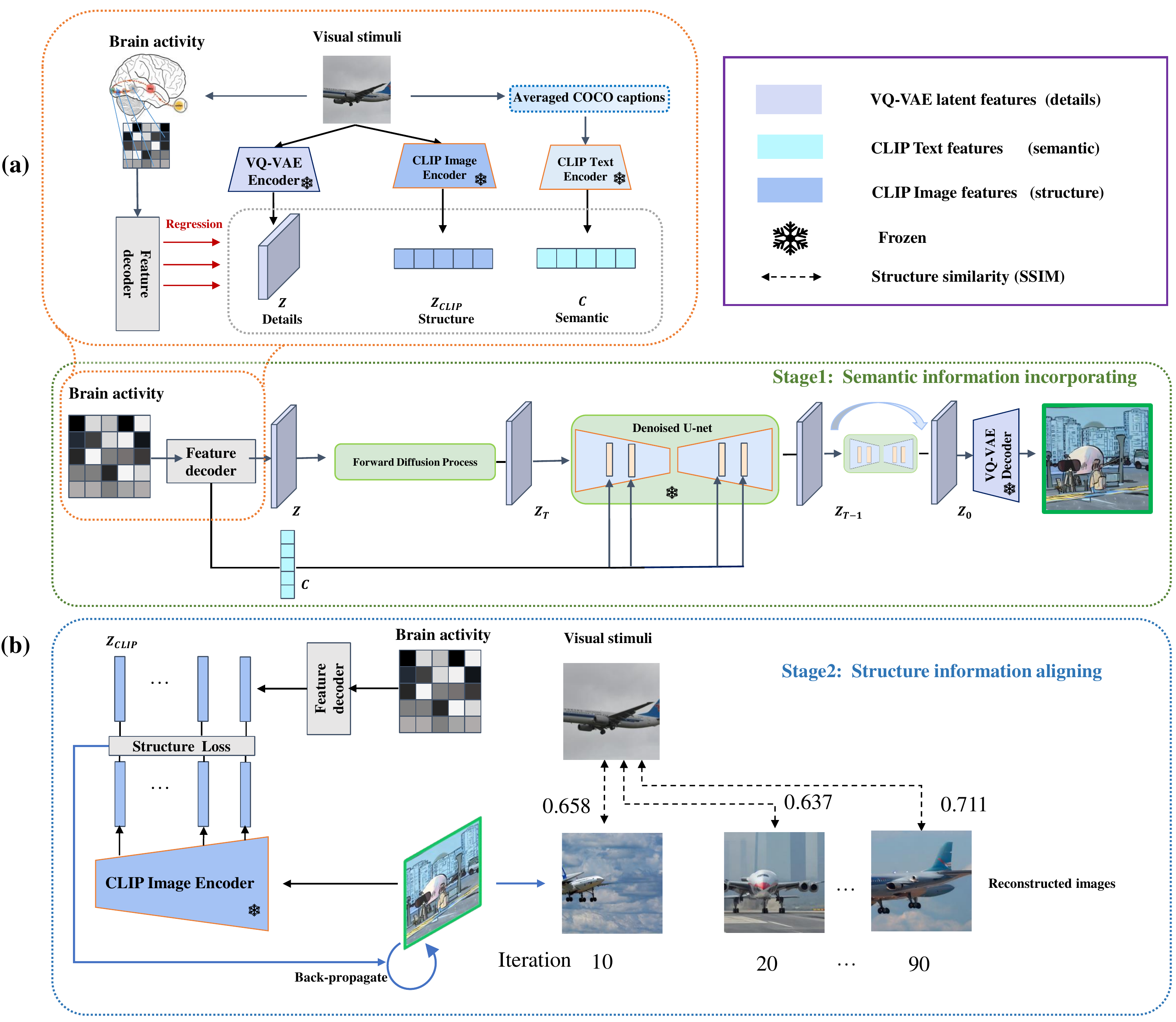}
	\caption{Schematic diagram of MindDiffuser. In Figure \ref{fig:overview}a, decoders are trained to fit fMRI with averaged CLIP text embeddings $c$, CLIP image features $Z_{CLIP}^i$, and VQ-VAE latent features $z$. Figure \ref{fig:overview}b illustrates the two-stage image reconstruction process. In stage 1, an initial reconstruction image is generated using the decoded CLIP text feature $c$ and VQ-VAE latent feature $z$. In stage 2, the decoded CLIP image features are used as a constraint to iteratively adjust $c$ and $z$ until the final reconstruction result matches the original image in terms of both semantics and structures.}
	\label{fig:overview}
\end{figure}

In this section, we present {\bfseries MindDiffuser}, a two-stage model for controlled image reconstruction. Briefly, in stage 1, we decode fMRI into CLIP text embeddings $c$ and visual features $z$ in the VQ-VAE latent space. This enables the initial reconstructed image generated by Stable Diffusion to contain semantic information and fine-grained details, thereby interpreting {\bfseries ``What is contained in the image ?''} In stage 2, we decode the fMRI signal into the low-level visual features of CLIP, and continuously optimize $c$ and $z$ in stage 1 by back-propagation, so that the generated image approximates the ground truth in the CLIP visual feature space, thus achieving control over the structural information and deciphering {\bfseries ``Where are the objects in the image ?''}

Although stage 1 of our proposed model shares similarities with the approach of Takagi et al.\cite{takagi2022high}, our primary objective is to enhance the controllability of image reconstruction. Our proposed model aims to ensure that the reconstructed images not only exhibit semantic similarity but also maintain structural alignment with the original image.

\subsection{Stage 1}
{\bfseries Semantic information incorporating --- What is contained in the image ?}

Suppose that $X \in R^{D_x \times N}$ and $Y \in R^{D_y \times N}$ denote the fMRI activity patterns and its corresponding visual stimuli in the training set, respectively. And $c \in R^{D_c \times N}$, $z \in R^{D_z \times N}$ and $Z_{CLIP}^i \in R^{D_i \times N}$ denote CLIP text embeddings, VQ-VAE latent vectors and the visual features of layer i in CLIP extracted from the training set. Here, $D_j$ denotes the dimentions of aforementioned data,  and N denotes the size of the training set.
Figure \ref{fig:overview} (a) illustrates the training process of three linear regression models: $f_c:X \mapsto c$ , $f_z:X \mapsto z$ and $f_{CLIPi}:X \mapsto Z_{CLIP}^i$ using the training dataset. The trained $f_c$ and $f_z$ are utilized to decode the CLIP text embedding $c$ and latent vector $z$ of the image that needs to be reconstructed. Subsequently, these two features are fed into the image-to-image process of Stable Diffusion, as illustrated in Figure \ref{fig:overview} (b). Specifically, decoded $z$ undergoes a forward diffusion process, as outlined by equations \ref{equ1} and \ref{equ2} , resulting in the computation of $z_T$.
\begin{equation}\label{equ1}
	q(z_t|z_{t-1})=\mathcal{N}(z_t;\sqrt{\alpha_t}z_{t-1},(1-\alpha_t)I)    \quad t=0,1,\cdots T 
\end{equation}
\begin{equation}\label{equ2}
	z_T=\sqrt{\overline{\alpha_T}} z+ \sqrt{1-\overline{\alpha_T}}\epsilon        \quad  and \quad z_0=z
\end{equation}

In each reverse denoising iteration, the U-Net integrates the decoded CLIP text embedding $c$ into $z_T$ using cross-attention, as defined in Equation \ref{equ3}.
\begin{equation}\label{equ3}
	CrossAttention(Q,K,V)=softmax(\frac{Q K^T}{\sqrt{d}}) \  \  Q=W_Q^i\cdot\phi_i(z_t), K=W_K^i\cdot c, V=W_V^i\cdot c
\end{equation}
where $\phi_i(z_t)$ represents the U-Net middle-layer feature, $c$ corresponds to the decoded CLIP text information, and $W_Q^i$, $W_K^i$, $W_V^i$ denote the pre-trained projection matrixs. In this way, we reformulate the optimization objective of Stable Diffusion to Equation \ref{equ4}.
\begin{equation}\label{equ4}
	L_{Semantic}^t= \mathbb{E} _{z_t,\epsilon\sim\mathcal{N}(0,1),t}[\Vert\epsilon-\epsilon_\theta(z_t,t,c,\phi_i(z_t)) \Vert_2^2]
\end{equation}
where $\epsilon_\theta( \cdot )$ is a set of denoising functions that are usually
implemented as U-Net. The images generated through this process contain semantic information and fine-grained details. And the decoded $z$ first undergoes a forward diffusion process, which introduces some degree of variability in the resulting images.

\subsection{Stage 2}
{\bfseries Structure information aligning --- Where are the objects in the image ?}

In stage 1, we employ the decoded CLIP text embedding $c$, and the VQ-VAE latent vector $z$ to generate an initial reconstructed image $\hat{Y}$ that contains semantic information. The CLIP visual branch, denoted as $\Phi$, encodes high-level semantic information at the last layer and lower-level structural information such as posture and position at the shallow layers. To align the structural information of the reconstructed image with that of the original image, without losing the semantic information, we extract the shallow linear layer features of the CLIP image encoder, as illustrated in Figure \ref{fig:overview} (b). Then, we decode the CLIP visual features corresponding to each layer using fMRI, and calculate the L2 distance between the two:
\begin{equation}\label{equ5}
	L_{Structure}=\sum\limits_{i=1}\Vert \Phi^i_{CLIP}(\hat{Y})-Z^i_{CLIP} \Vert_2^2
\end{equation}
Subsequently, the decoded $c$ and $z$ from the first step are continually updated through backpropagation, to achieve the goal of controlling the reconstructed output.

\section{Experiments}
\subsection{Dataset}
NSD\cite{allen2022massive} is currently the largest neuroimaging dataset bridging brain and artificial intelligence, consisting of densely sampled fMRI data from 8 subjects. Each subject viewed 9000-10000 different natural scenes (with 22000-30000 repetitions) during 30-40 MRI scanning sessions, using whole-brain gradient-echo EPI at 1.8 mm isotropic resolution and 1.6 s TR for 7T scanning. The image stimuli viewed by the subjects are obtained from the Common Objects in Context (COCO)\cite{lin2014microsoft} dataset and corresponding captions can be extracted using the COCO ID of the image stimuli.

To validate the stability of MindDiffuser across different subjects, we conducted experiments using fMRI-stimulus pairs from subjects 1, 2, 5, and 7 in NSD. The training set for each subject contains 8859 image stimuli and 24980 fMRI trials (with each image having up to 3 trials). Additionally, 982 image stimuli and 2770 fMRI trials are shared among the four subjects. For fMRI data with multiple trials, we computed the average response. The properties of the dataset used in our experiments have been summarized in Table \ref{tab:my-table}.

\begin{table}[htbp!]
	\renewcommand\arraystretch{1.2}
	\centering
	\begin{tabular}{c|c|c|c|c|c}
		\toprule
		Dataset & Training & Test(shared) & ROIS & Subject ID & Voxels \\ \hline
		\multirow{4}{*}{NSD} & \multirow{4}{*}{8859} & \multirow{4}{*}{982} & \multirow{4}{*}{\begin{tabular}[c]{@{}c@{}}V1, V2, V3, V4,\\ PHC, MT, MST, LO, IPS\end{tabular}} & sub01 & 11694 \\ 
		&          &              &      & sub02      & 9987   \\ 
		&          &              &      & sub05      & 9312   \\ 
		&          &              &      & sub07      & 8980   \\ \bottomrule
	\end{tabular}
\caption{The details of NSD used in our experiments.}
\label{tab:my-table}
\end{table}

\subsection{Feature decoding experiments}
During the image reconstruction process, we first utilize the \href{https://github.com/KamitaniLab/PyFastL2LiR}{FastL2LiR} , developed by KamitaniLab, to train three distinct linear regression models, which decode fMRI data into three distinct spaces:

{\bfseries CLIP text feature space}

The Stable Diffusion utilizes a CLIP text encoder with a feature space dimension of 77$\times$768, where 77 denotes the maximum token length and 768 represents the encoding dimension of each token. To account for the typical caption lengths in the COCO dataset, which rarely exceed 15 tokens, the first 15$\times$768 dimensions of the flattened features are used during practical operation. The features in this space inject {\bfseries semantic } information into the reconstructed images.

{\bfseries VQ-VAE latent space}

To integrate richer {\bfseries details}, we extract latent space features (1$\times$4$\times$64$\times$64 dimensions) of the images in training set using VQ-VAE (included with Stable Diffusion) and flatten them. Subsequently, fMRI signals are mapped to this latent space. Results of experiments conducted in the Figure \ref{fig:ablation} underscore the importance of incorporating fine-grained details in image reconstruction to accurately align the structural information with the original images.

{\bfseries CLIP image feature space}

In order to align the {\bfseries structural} information of the reconstructed images with corresponding ground truth without losing semantic information, we choose the low-level visual features of CLIP to control the reconstructed image, as shown in Figure \ref{fig:overview} . We choose ViT/B-32 as the backbone for pre-trained CLIP, and extract features from several layers (Linear-2, Linear-4, $\cdots$ Linear-12) in the CLIP vision branch. These layers have 38400 dimensions each. Due to the potential impact of those low-accuracy features on the decoding process, we first calculate the predictive accuracy (Pearson correlation coefficient) of the CLIP features for each dimension on the training set using a 5-fold cross-validation before formally fitting each layer of features. We then select the top 25\% of features based on their predictive accuracy and re-fit these features using all the training data. During the reconstruction of images on the test set, only these 25\% of features are used to guide the alignment of the original and reconstructed images.

\subsection{ Image reconstruction from decoded features}

\subsubsection{Overview of our reconstruction results}
\begin{figure}[htbp!]
	\centering
	\includegraphics[width=1.0\linewidth]{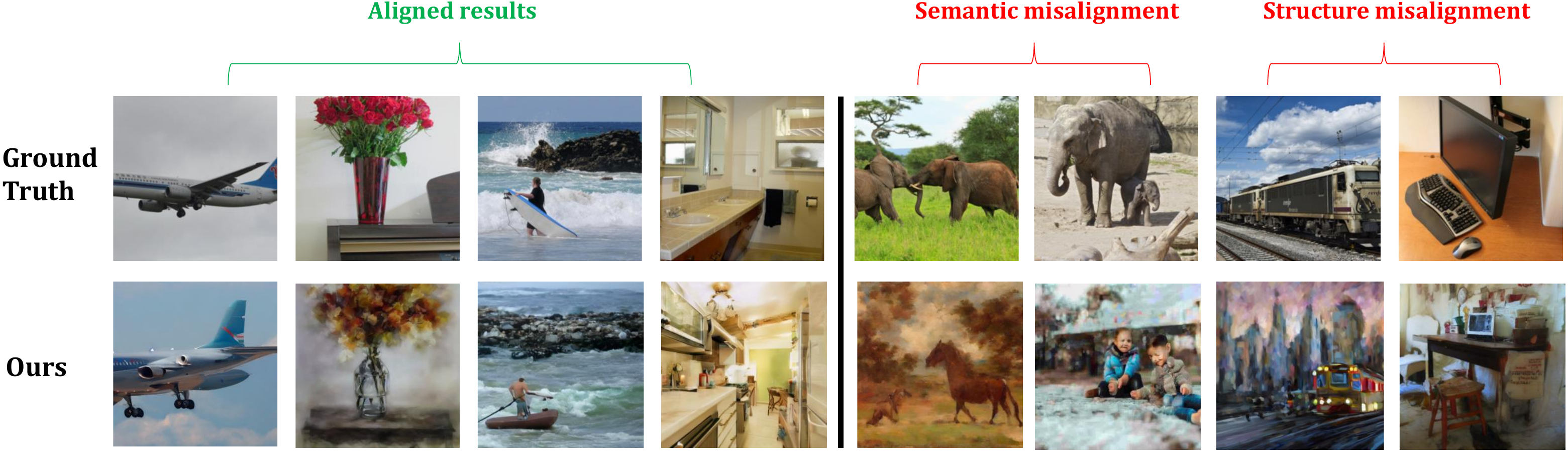}
	\caption{The image reconstruction results of MindDiffuser. The left side of the black line shows results with good alignment of both semantic and structural information to the original image. The right side displays results with misalignment of semantic or structural information.}
	\label{fig:result-overview}
\end{figure}

Following the process in Figure \ref{fig:overview} , some of the images reconstructed by MindDiffuser are shown in Figure \ref{fig:result-overview} . (A detailed reconstruction results can be found in the \cref{ssc:ap_data}.)
As depicted in the left side of Figure \ref{fig:result-overview} , precise decoding of both semantic and structural guidance information leads to near-perfect reconstruction of the original images, achieving the controllable reconstruction of corresponding image stimuli based on fMRI. However, when only the structural information is decoded correctly,  the reconstructed images differ significantly from the original images. For example, the original images of two elephants may be reconstructed as two horses or two children. Although the reconstructed image may be aligned with the original in terms of position and background, the semantic information is lost. Similarly, if the semantic information is decoded accurately while the structural information is not, as shown below, it may be possible to generate images of a train and a laptop, but their orientation, position, and size may not be aligned with the original images.
\subsubsection{Comparison with state-of-the-art models}
\begin{figure}[htbp!]
	\centering
	\includegraphics[width=1.0\linewidth]{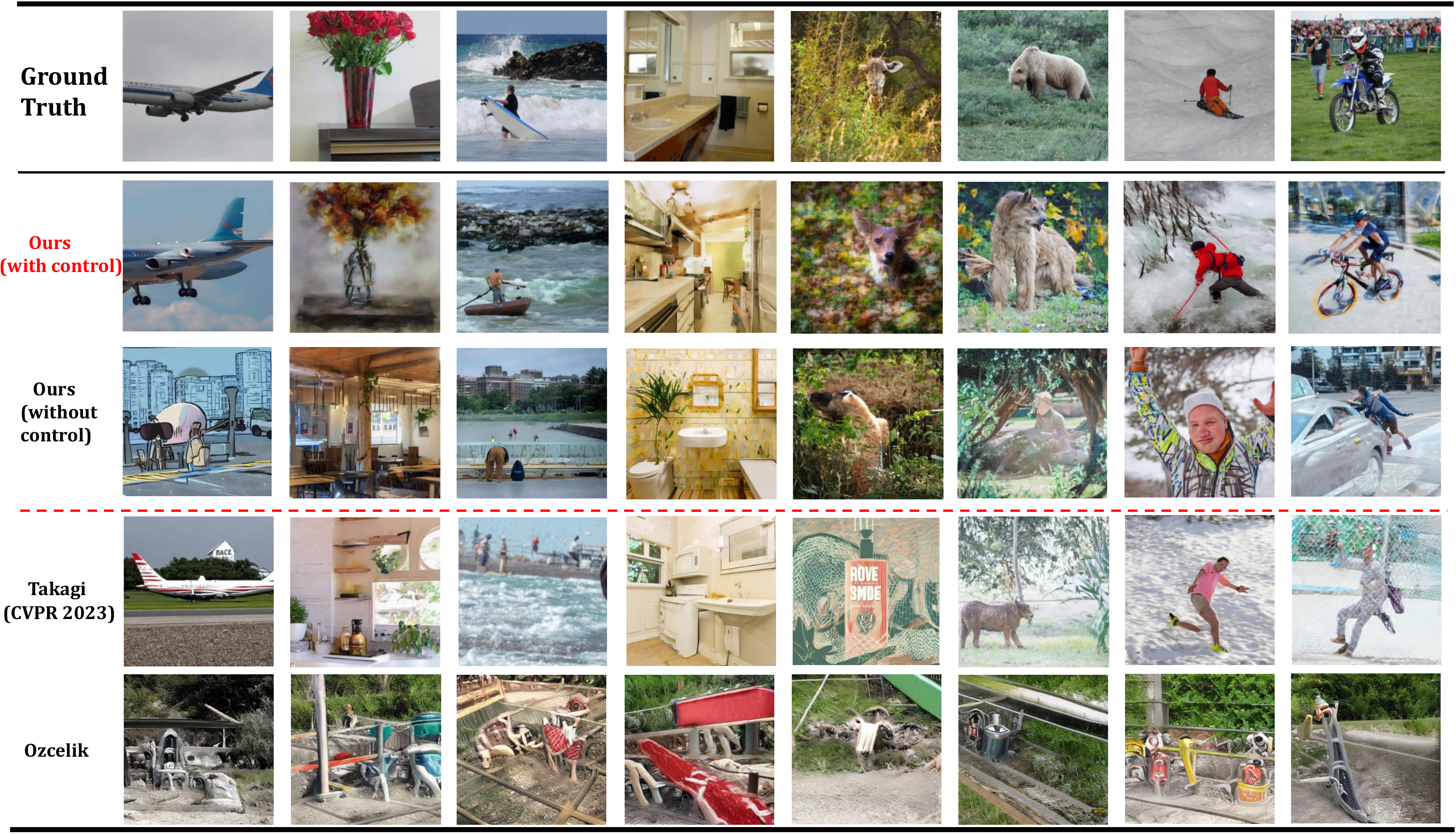}
	\caption{A comparative analysis of reconstruction models on NSD. The first line represents the stimulus images. The second and third lines represent the results obtained using our proposed MindDiffuser with and without stage 2, respectively. The fourth and fifth lines represent the results obtained by reproducing the experiment described in Takagi's \cite{takagi2022high} paper and by using Ozcelik's\cite{ozcelik2022reconstruction} provided random version code, respectively.}
	\label{fig:compareall}
\end{figure}

In order to effectively align the reconstructed outputs with the original images in terms of both semantic and structure, it is essential to ensure a high decoding accuracy for both. To achieve this, we select test set images with a decoding accuracy of 0.3 or higher, focusing specifically on the preservation of semantic and structural information. As much of the existing work on image reconstruction utilizing NSD has not yet been made fully open source, we took it upon ourselves to reproduce the results of Takagi et al.\cite{takagi2022high} and Ozcelik et al.\cite{ozcelik2022reconstruction} on NSD, in order to facilitate direct comparison with our findings, as shown in Figure \ref{fig:compareall} .
According to Figure \ref{fig:compareall}, it can be observed that compared to recent work, our approach produces reconstructed results on NSD that are visually more similar to the ground truth both in terms of semantic and structure. This suggests that our approach outperforms existing methods in controllable image reconstruction on NSD. We only have access to the random version code released by Ozcelik\cite{ozcelik2022reconstruction}, and based on their paper, the quantitative results between the random and dense versions are similar. However, this model did not perform well on NSD. We speculate that this could be due to the fact that NSD contains more complex natural scenes compared to the previous Generic Object Decoding ( \href{https://figshare.com/articles/dataset/Generic_Object_Decoding/7387130/6}{GOD} ) dataset\cite{horikawa2017generic}. Extracted instance features using only self-supervised pre-trained feature extractor SWAV\cite{caron2020unsupervised} on the visual modality may not contain complete semantic and structural information, leading to unsatisfactory results even if the decoding accuracy of these features is high (with an average accuracy of around 0.6 on the test set). It has been observed that although Takagi et al.\cite{takagi2022high} and MindDiffuser without stage 2 are able to achieve semantically similar reconstruction results to the original images, such as airplanes, plants on a table, people surfing at a beach, bathrooms, and people skiing, it is clear that these images cannot be aligned with the original images in terms of details such as shape and orientation. This indicates that the CLIP-based modulation used in stage 2 of this paper can effectively reconstruct results that are both semantically and structurally aligned with the original images.

\begin{table}[]
	\centering
	\renewcommand\arraystretch{1.2}
	\setlength{\tabcolsep}{7mm}{
	\begin{tabular}{c|c|cc}
		\toprule
		\multirow{2}{*}{Methods} & Semantic Similarity {\bfseries $\uparrow$} & \multicolumn{2}{c}{Structure Similarity {\bfseries $\uparrow$} } \\ \cline{2-4} 
		& CLIP                & SSIM                & PCC                \\ \hline
		\multirow{2}{*}{\begin{tabular}[c]{@{}c@{}}Ozcelik\cite{ozcelik2022reconstruction} \\ Takagi(CVPR2023)\cite{takagi2022high} \end{tabular}} &
		\multirow{2}{*}{\begin{tabular}[c]{@{}c@{}}0.546\\ 0.642\end{tabular}} &
		\multirow{2}{*}{\begin{tabular}[c]{@{}c@{}}0.135\\ 0.300\end{tabular}} &
		\multirow{2}{*}{\begin{tabular}[c]{@{}c@{}}0.126\\ 0.175\end{tabular}} \\
		&                     &                     &                    \\ \hline
		\multirow{3}{*}{\begin{tabular}[c]{@{}c@{}}Ours(without control)\\ Ours(without z)\\ Ours(full model)\end{tabular}} &
		\multirow{3}{*}{\begin{tabular}[c]{@{}c@{}}0.597\\ 0.616\\{\bfseries0.765}\end{tabular}} &
		\multirow{3}{*}{\begin{tabular}[c]{@{}c@{}}0.253\\ 0.292\\ {\bfseries 0.354}\end{tabular}} &
		\multirow{3}{*}{\begin{tabular}[c]{@{}c@{}}0.183\\ 0.066\\ {\bfseries0.278}\end{tabular}} \\
		&                     &                     &                    \\
		&                     &                     &                    \\ \bottomrule
	\end{tabular}     }
	\caption{Quantitative comparison of image reconstruction.}
	\label{tab:table_2}
\end{table}

\begin{figure}[htbp!]
	\centering
	\includegraphics[width=1.0\linewidth]{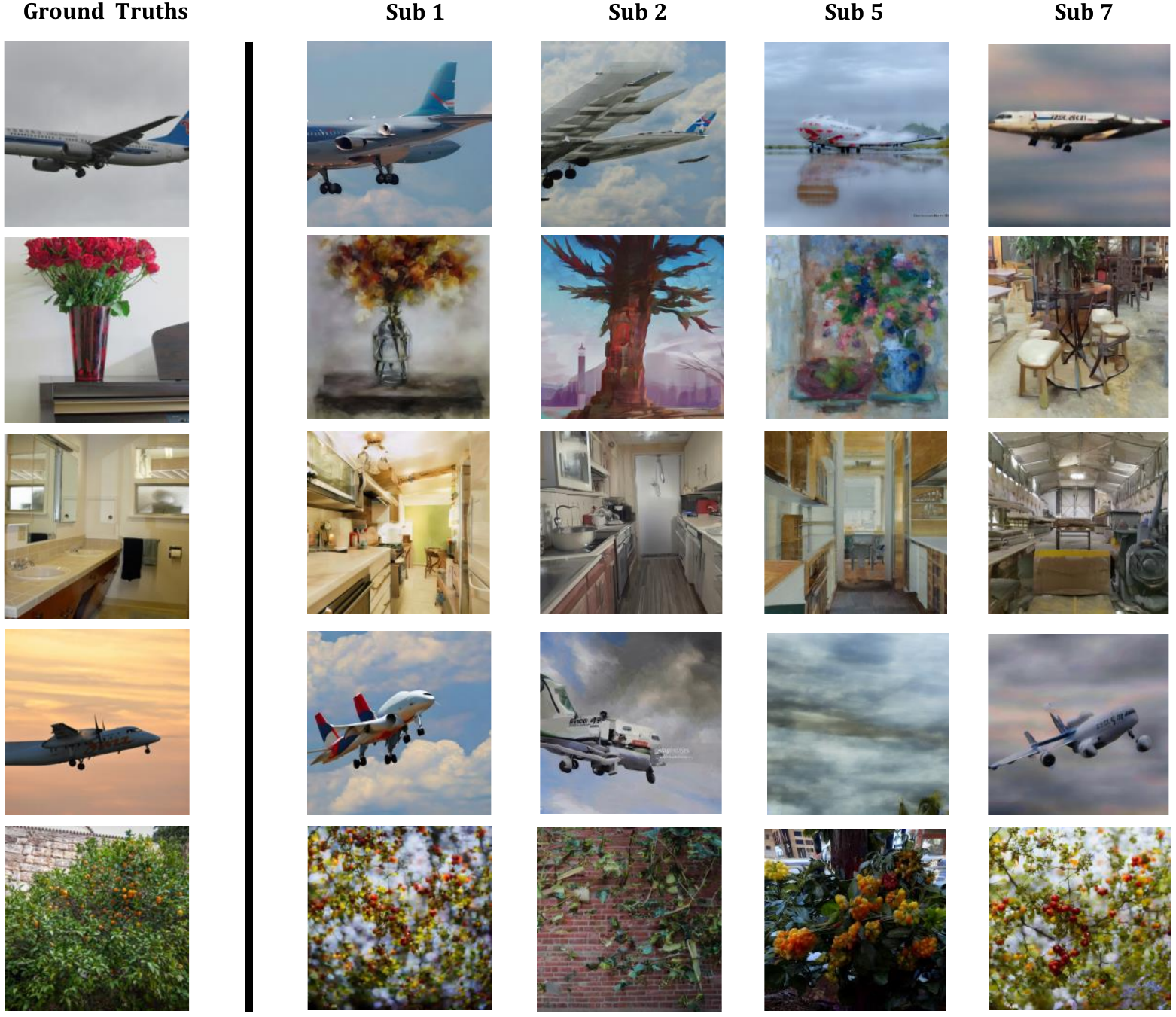}
	\caption{Reconstruction results of MindDiffuser on multiple subjects}
	\label{fig:foursub}
\end{figure}

\begin{figure}[b!]
	\centering
	\includegraphics[width=1.0\linewidth]{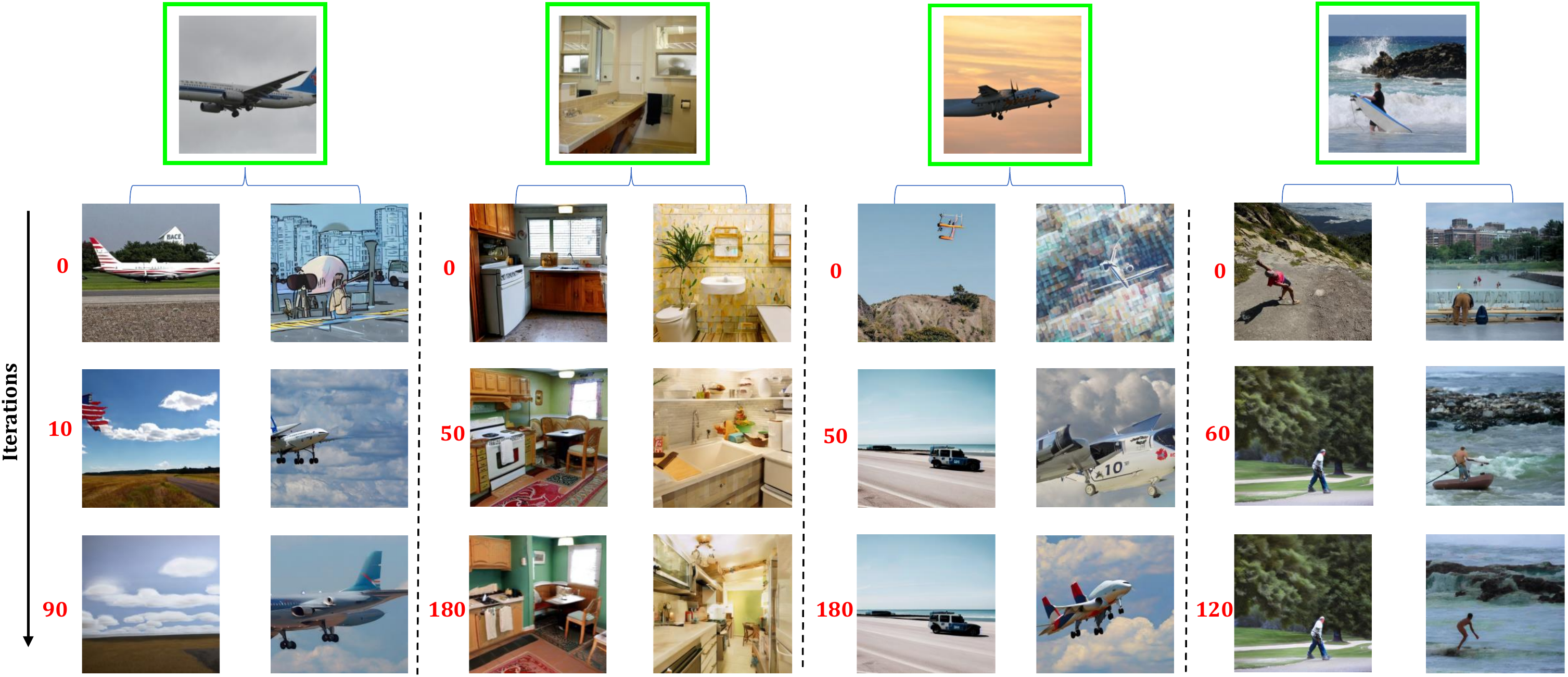}
	\caption{Results of ablation experiments on details $z$. The ground truth is marked with a green box. For each image to be reconstructed, the left column shows the multi-step iterative results obtained by decoding only $c$ and CLIP visual features, and then randomly sampling $z$ from the latent space of VQ-VAE. The right column shows the multi-step iterative results obtained by our proposed complete model.}
	\label{fig:ablation}
\end{figure}

In order to further quantitatively compare the reconstruction performance of our method with some state-of-the-art methods, we utilize three evaluation metrics to compare from both semantic and structural aspects. We use cosine similarity in the last layer of CLIP visual branch (512 dimensions) to measure the semantic similarity between the reconstructed results and the original images. We use SSIM and per-pixel correlation coefficient to measure the structural similarity between them. All the three metrics range from 0 to 1, and higher values indicate better reconstruction results. The results from Table \ref{tab:table_2} demonstrate that our reconstruction surpasses the current state-of-the-art models both semantically and structurally. Furthermore, we conducted two ablation experiments to confirm the importance of both the fine-grained details $z$ and the CLIP supervision in Stage 2 of our model. Please refer to Figures \ref{fig:compareall} and \ref{fig:ablation} for qualitative details.
\subsubsection{Adaptability across 4 subjects}
The anatomical structure and functional connectivity of the brain vary among individuals, resulting in differences in the fMRI signals even when the same image stimulus is presented. To validate our proposed MindDiffuser's ability to adapt to inter-subject variability, we reconstruct the test images of subjects 1, 2, 5, and 7 without any additional adjustments. The results are shown in Figure \ref{fig:foursub}.

As depicted in Figure \ref{fig:foursub}, the identical image stimulus may result in different and unsatisfactory reconstruction results for some subjects due to differences in the subjective brain responses during fMRI acquisition and disparities in the accuracy of feature decoding. For instance, "flowers on the table" is erroneously reconstructed as "table and chair" in subject 2, and "aircraft at sunset" can not be reconstructed in subject 5. However, for the majority of the reconstructed images, our model achieve a satisfactory alignment with the original images in terms of both semantic and structural features for each selected subject, underscoring the ability of MindDiffuser to effectively accommodate variations among individuals.

\subsubsection{The importance of fine-grained details $\textbf{z}$}

In the context of generative image reconstruction models, Lin\cite{lin2022mind} and Chen\cite{chen2022seeing} leverage decoded semantic information as a guiding condition, and randomly sample detailed information from the latent space of a conditional generative model such as StyleGAN-2 or LDM to reconstruct images. Notably, Takagi's\cite{takagi2022high} work demonstrate that even without adequate details, decoding semantic information alone can produce results that are comparable to those of complete models. Besides, as representative of optimization-based models, Shen\cite{shen2019deep} sample details $z$ randomly from the latent space of a GAN, which are subsequently optimized using DNN-extracted features to obtain reconstructed results. These studies suggest that high-fidelity semantic information or structural information from DNN employed as supervisory signals can result in satisfactory reconstruction outcomes, even in the absence of precise details. In this regard, we will conduct experimental investigations to evaluate the importance of fine-grained details $z$ in our proposed model.

Figure \ref{fig:ablation} illustrates that the provision of precisely decoded details $z$ to the model, coupled with the guidance of CLIP visual features, engenders a reconstruction that is semantically and structurally aligned with the original image, following several iterations. Conversely, if the details $z$ are initialized randomly, the model is rapidly enmeshed in a suboptimal solution during the optimization process, resulting in unsatisfactory outcomes. Specifically, regarding the reconstruction of the second image, "bathroom", after 180 iterations, the resultant image reflects the original's line direction but exhibits substantial discrepancies in terms of color and semantics. Similarly, the reconstruction of the third image, "airplane", after 180 iterations, produces an image of a car instead, while also a vehicle but belongs to a distinct category. These findings corroborate that fine-grained details $z$ not only govern the reconstructed images' size, orientation, and texture, but also constitute a complementary  adjunct to semantic information $c$, facilitating smooth progression of the optimization process in stage 2.

\section{Conclusion}

We propose a two-stage image reconstruction model, {\bfseries MindDiffuser}, which aligns both semantic and structural information of the reconstructed images with the image stimuli on NSD. In stage 1, we decode fMRI into CLIP text feature $c$ and VQ-VAE latent feature $z$, and apply forward noise injection to $z$ to obtain $z_T$. Then, we fuse $z_T$ and $c$ through cross-attention in the backward denoising process to generate an initial image with semantic information and fine-grained details. In stage 2, we use the CLIP low-level visual features decoded from fMRI as supervisory signals to continuously adjust the two features in stage 1 through backpropagation, aligning the reconstructed image's structural information with that of the original image. MindDiffuser outperforms some state-of-the-art models qualitatively and quantitatively on NSD, given the high decoding accuracy of stage 1. Moreover, our experiments show that MindDiffuser is adaptive to inter-subject variability, achieving excellent reconstruction results for the image stimuli of subjects 1, 2, 5, and 7 without any additional adjustment.

\bibliographystyle{unsrt}
\bibliography{reference}

~\\
~\\
\hspace*{\fill} \ 


\appendix

\section{Appendix}

\subsection{More Reconstruction Samples}
\label{ssc:ap_data}
More reconstructed images of MindDiffuser on subject 1 are shown as follows:
\begin{figure}[htbp!]
	\centering
	\includegraphics[width=1.0\linewidth]{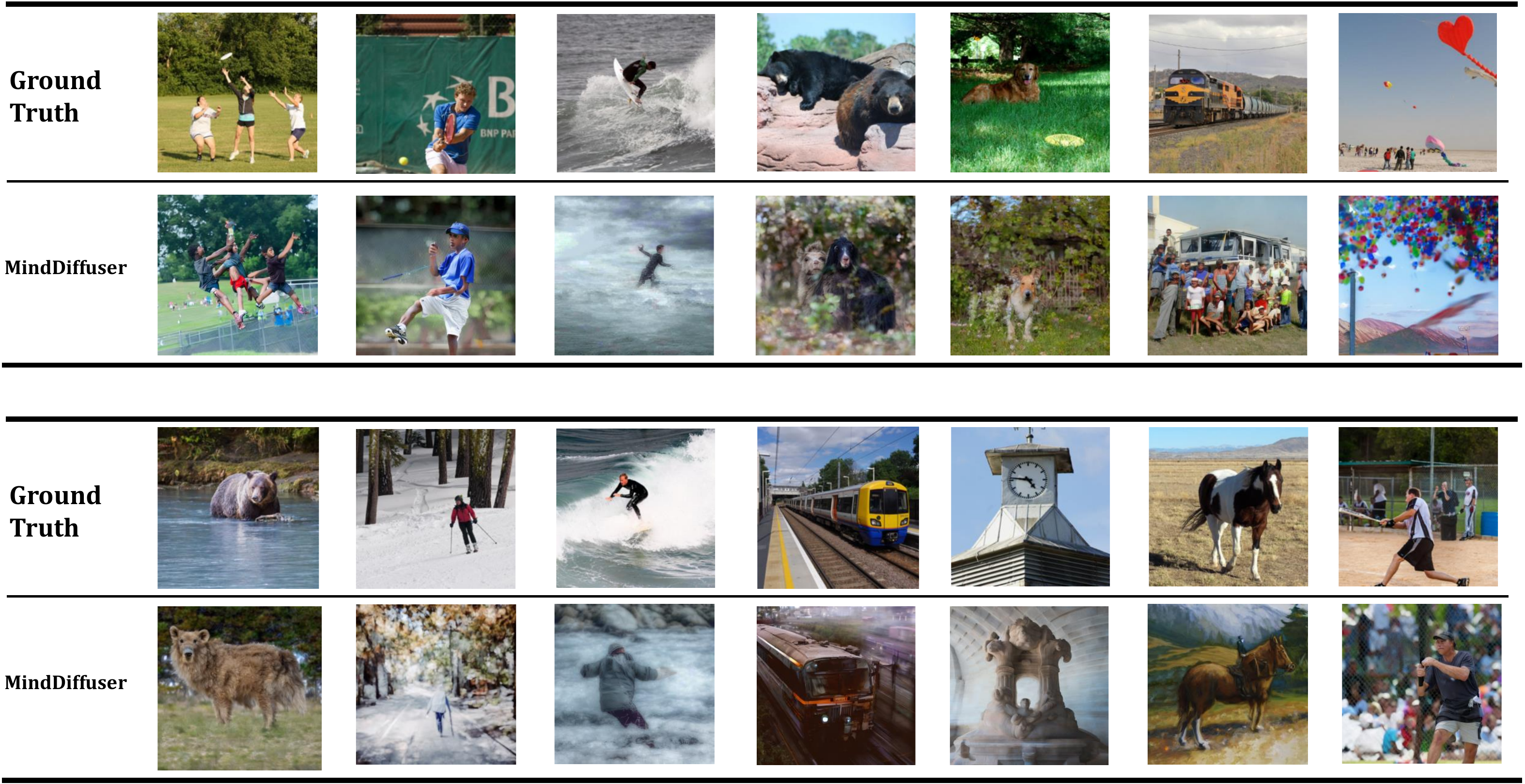}
	\label{fig:appendixpicture1}
\end{figure}
\begin{figure}[htbp!]
	\centering
	\includegraphics[width=1.0\linewidth]{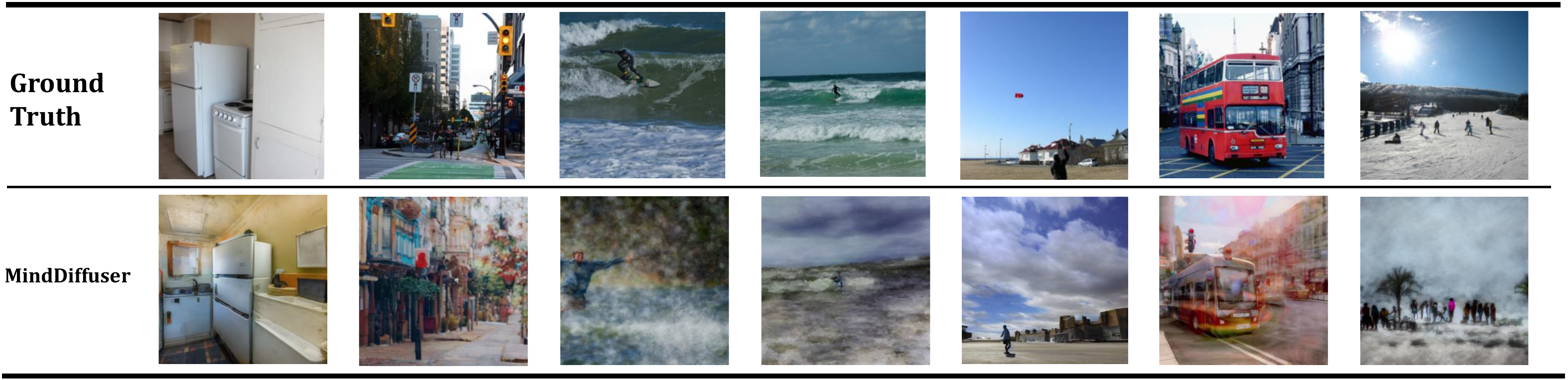}
	\caption{More samples for sub1 on NSD}
	\label{fig:appendixpicture2}
\end{figure}



\end{document}